\documentclass{article}
\usepackage{iclr2026_conference,times}
\usepackage[T1]{fontenc}
\usepackage[utf8]{inputenc}
\usepackage{amsmath,amssymb,graphicx,booktabs,array,tabularx}
\usepackage{hyperref,url}
\iclrfinalcopy
\hypersetup{hidelinks,pdftitle={CVIF: Steering Geometric Evidence During Prediction Formation},pdfauthor={Anonymous}}

\usepackage{multirow}

\title{
\parbox{\textwidth}{
\centering
\normalfont\Large\bfseries
CVIF: A Criticality-Driven Visual Intervention Framework\\
for Geometric Diagram Understanding in MLLMs
}
}

\author{
\begin{tabular}{c}
Jiahui Kang$^{1}$, Bifan Wei$^{1,\dagger}$,
Lingling Zhang$^{1}$, Tianwen Jiang$^{2}$ \\[2pt]
Qiuyong Xiao$^{2}$, Jihong Zhang$^{2,\dagger}$,
Jun Liu$^{1}$ \\[4pt]
\normalfont
$^{1}$School of Computer Science and Technology, Xi'an Jiaotong University
$^{2}$Tencent Hy AI Data \\[2pt]
\normalfont\small
jhkang@stu.xjtu.edu.cn
$^{\dagger}$Corresponding authors
\end{tabular}
}

\begin{document}
\maketitle

\begin{abstract}
Despite significant progress in visual tasks by Multimodal Large Language Models (MLLMs), geometric diagram understanding remains challenging due to the presence of sparse visual cues and ambiguous symbol-primitive associations. MLLMs may therefore rely on textual priors, producing interpretations that conflict with visual evidence. We introduce the training-free Criticality-Driven Visual Intervention Framework (CVIF), an inference-time method that localizes critical layers and executes visual interventions during the transition from evidence aggregation to semantic decoding. At these layers, a Geometry-Constrained Local Relation Reconstruction (GCLR) module selects and weights vertex-centered visual evidence, while an Adaptive Visual Steering Operator (AVSO) redistributes attention mass toward the selected tokens.
Experiments on PGPS9K and PGDP5K show that CVIF raises Overall F1 from 77.85 to 85.58 and from 75.23 to 82.84, respectively, establishing a novel inference-time visual intervention paradigm.
\end{abstract}

\section{Introduction}
In recent years, Multimodal Large Language Models (MLLMs) have achieved substantial progress in the field of multimodal understanding \citep{lu2024mathvista, wang2024mathvision, zhang2025open, qiao2025wemath, chen2026mathflow}.
Geometric Problem Solving (GPS), which integrates geometric diagrams with textual questions, has emerged as a crucial benchmark for evaluating their cognitive capabilities \citep{ma2026survey, chen2021geoqa}. However, existing MLLMs tend to deduce visual information for geometric reasoning from textual contexts rather than genuinely perceiving the diagrams. Extensive empirical studies reveal that the current end-to-end GPS paradigm \citep{zhang2024geoeval, zhang2024mathverse} conflates perception errors with reasoning failures, making it difficult to isolate the true source of model limitations \citep{sun2026math, zhang2024mathverse, guo2025large, wang2026geoparsing, cho2025geodano}.
Multiple diagnostic studies consistently confirm that perception errors remain the primary cause of failures in geometric reasoning tasks \citep{wang2025mvmath, wang2025solidgeo} (Figure 1(a)). Therefore, it is imperative to explicitly decouple visual perception from logical reasoning by requiring models to independently parse geometric diagrams and generate formal language descriptions, thereby enabling an objective and unambiguous evaluation.

\begin{figure}[htbp]
    \centering
    \includegraphics[width=0.98\linewidth]{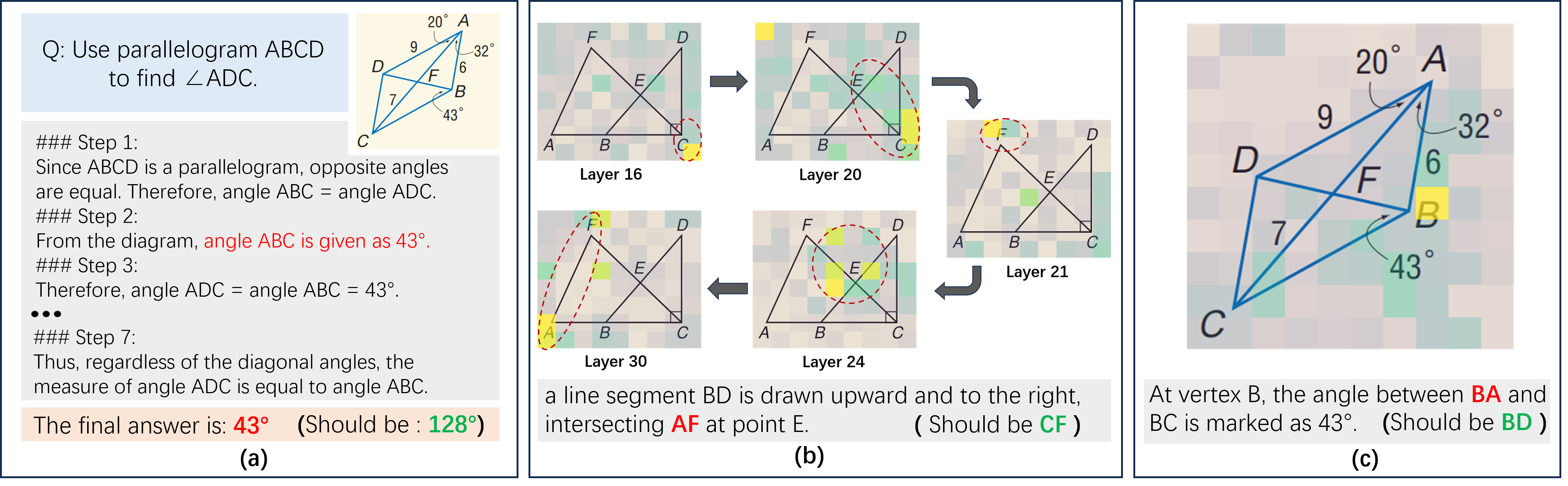}
    \caption{Attention failure modes of MLLMs in geometric diagram perception.
        (a) Perceptual errors during the process of geometric reasoning lead to prediction failures.
        (b) When predicting which line segment intersects $BD$ at point $E$, shallow layers attend to the visually correct point $C$. As depth increases, attention shifts toward evidence that confirms a prior-compatible candidate (e.g., $AF$); by a later layer, the visually correct point $C$ receives little attention.
        (c) When determining which angle is annotated by $43^\circ$, late-layer attention may concentrate on the label ``B'' and nearby symbols while failing to recover the local topology of the three edges incident to B. In particular, the model does not sufficiently use edge $BD$ as structural evidence for resolving the angle assignment.}
\end{figure}

Why do MLLMs readily make errors in perceiving geometric diagrams? Our study finds that these errors can be traced to: (1) visual cues captured at early layers being overlooked in deeper layers, and (2) attention mechanisms in deeper layers becoming excessively focused on letter symbols while overlooking geometric structures. This is precisely consistent with recent observations from extensive research on MLLM vision that such failures are primarily not caused by an inability to perceive visual information \citep{khayatkhoei2025mllms,liu2026seeing,shao2026thinking}.
First, attending to relevant visual tokens does not ensure their continued use in later layers. As illustrated in Figure 1(b), textual priors can favor an incorrect interpretation of sparse visual inputs \citep{favero2024multi,liu2024paying}. Subsequent attention updates may reinforce features consistent with that interpretation while reducing the contribution of conflicting visual evidence. We refer to this reinforcement of an existing interpretation as confirmation bias in the following discussion.
Second, in geometric tasks, attending to the correct symbols does not by itself establish the geometric relations between them. As shown in Figure 1(c), recognizing a number or a vertex label is insufficient to determine which incident edges define the annotated angle.
Human perception combines prior knowledge with incoming sensory evidence. When the two conflict, visual evidence can revise an initial expectation before a final judgment is formed. By contrast, an MLLM may increasingly favor a prior-consistent interpretation during autoregressive generation, even when the diagram contains conflicting visual cues. Once these cues lose influence in the model's internal representations, output-level correction alone cannot reliably restore them. This observation motivates intervention inside the model, rather than only through external prompting or post-hoc correction.

Building on the above analysis, we aim to design an active intervention mechanism that can identify the critical locations for visual intervention and provide relevant geometric cues.
First, conceptualizing Transformer inference as an energy-descent process, we can view the failure illustrated in Figure 1 as a consequence of hidden states failing to sufficiently leverage visual cues to establish the direction of energy descent. As a result, hidden states are overly misled by textual priors and gradually become trapped in a local-minimum state. We introduce the concept of representational criticality to describe a state in which a hidden representation has integrated diverse cues but has not yet formed a stable preference for a specific interpretation. Representations in such a state are particularly sensitive to changes in the relative contribution weights of visual evidence and textual priors, making them ideal locations for visual intervention.
Second, in geometric diagrams, geometric vertices not only serve as anchors for linguistic descriptions but also act as key nodes for constructing higher-level geometric topological structures. We can therefore use geometric vertices as entry points and combine them with the model’s attention history to identify relevant visual evidence that encodes local topological information. Building on this idea, we propose CVIF, a training-free framework that uses layer-wise prediction diagnostics to select intervention layers and steers attention toward vertex-guided visual features.
Our main contributions are:
(i) We propose a Criticality-Driven Visual Intervention Framework, which mitigates confirmation bias and visual hallucinations in MLLMs by applying mechanism-level perturbations at the criticality.
(ii) We design a Geometry-Constrained Local Relation Reconstruction (GCLR) mechanism, which reconstructs topological relations within the high-dimensional semantic space.
(iii) We introduce an Adaptive Visual Steering Operator (AVSO), which safely reshapes multi-head attention distributions, enabling the model to resist deep-layer textual inertia without disrupting native logical reasoning.

\section{Related Work}
\paragraph{Geometric Problem Solving and Geometry Diagram Parsing.}
Geometric Problem Solving has long served as a core benchmark for evaluating the complex cognitive and spatial reasoning capabilities of artificial intelligence \citep{lu2021inter, seo2014diagram}. Recently, MLLMs have demonstrated strong capabilities on various mathematical reasoning benchmarks \citep{lu2024mathvista, qiao2025wemath}. However, their performance on GPS tasks remains unsatisfactory \citep{xu2025geosense}. Most existing GPS works rely on end-to-end benchmarks, which tend to conflate perception errors with reasoning failures \citep{zhang2024geoeval, zhang2024mathverse}. In fact, recent studies have identified that MLLMs often suffer from severe perceptual impairments during long-sequence generation \citep{peng2026deeper}. For geometric diagrams, this challenge is further exacerbated by their extremely sparse and abstract topological features. Models may rely on textual priors when interpreting geometric diagrams, producing descriptions that are plausible from the text but inconsistent with the visual evidence. To address this, Geometry Diagram Parsing approaches attempt to translate diagrams into formal languages \citep{zhang2022plane, wang2026geoparsing} to decouple perception from reasoning.

\paragraph{Mechanistic Interpretability and Representation Dynamics.}
Mechanistic interpretability studies seek to characterize the computations implemented by neural networks\citep{elhage2021mathematical}. Layer-wise analyses identify depth-dependent linguistic representations\citep{tenney2019bert} and show that feed-forward networks (FFNs) can function as key-value memories whose outputs contribute to vocabulary-space predictions\citep{geva2021transformer,geva2022transformer}. \citet{lad2026remarkable} propose stages of inference involving contextual feature construction, prediction ensembling, and final distribution sharpening. These studies motivate a functional interpretation of changes across layers, rather than a uniform view of the forward pass. Connections between attention and modern Hopfield networks also provide an energy-based perspective on associative retrieval\citep{ramsauer2021hopfield}. In multimodal settings, studies report reliance on textual priors and limitations in visual grounding\citep{deng2025words,lee2025vlind,tong2024eyes}. Together, these findings motivate investigating where visual evidence can be reinforced during prediction formation.

\section{Analysis of Intervention Locations and Geometric Cues}

\subsection{Critical Locations for Visual Intervention}

\begin{figure}[htbp]
  \centering
  \includegraphics[width=0.98\linewidth]{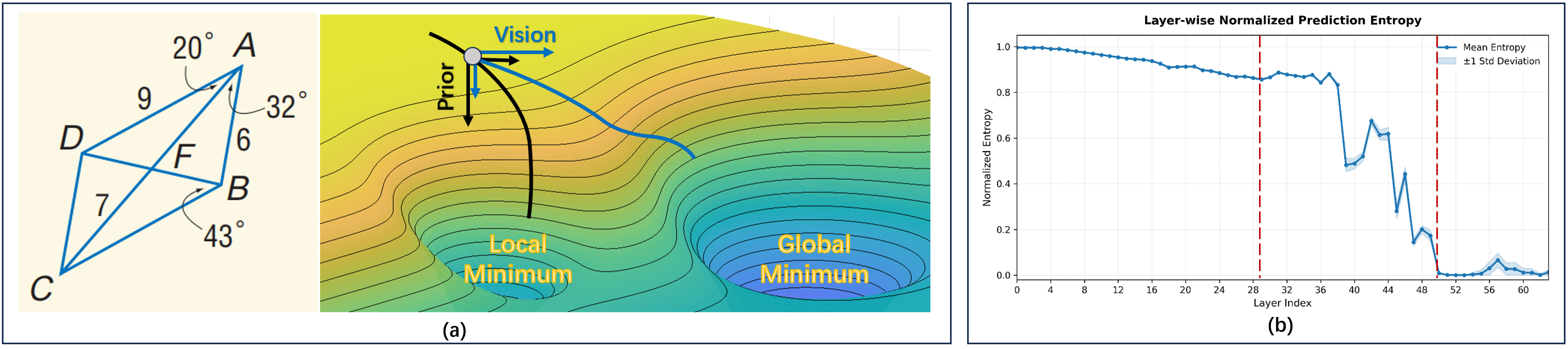}
  \caption{Prior-induced local minima and vision-guided trajectory correction.
      (a) When examining the angle at vertex B, textual priors tend to focus on the conventional $\angle$ ABC, whereas visual perception is naturally drawn to the salient feature "$43^\circ$".
      As depicted in the energy landscape, this causes the prediction state to be pulled by prior forces (black arrow), making it highly susceptible to being trapped in a neighboring local minimum that represents an erroneous prediction (i.e., $\angle ABC = 43^\circ$). During the critical state, introducing the local topological structure at vertex B provides a vision-driven steering force (blue arrow), thereby reducing the confidence of local minima induced by prior and steering the prediction toward the global minimum.
      (b) Layer-wise normalized predictive entropy for Qwen3-VL-32B at the question-end token.}
  \label{fig:landscape}
\end{figure}

To conceptualize the dynamics of forward inference, we adopt an energy-landscape perspective inspired by continuous Hopfield networks \citep{ramsauer2021hopfield}. From this viewpoint, we treat the forward pass as a process in which hidden states continually evolve toward stable states in a high-dimensional semantic space.
For geometric diagram understanding, interpretations supported by textual priors and those supported by local visual evidence may compete with each other (Figure~\ref{fig:landscape}(a)).
At representational criticality, small changes in the relative contributions of visual evidence and textual priors can alter the subsequent evolution of hidden states. Representational criticality therefore identifies a critical location for visual intervention: if intervention occurs too early, the model has not yet formed a semantic judgment relevant to the current prediction; if it occurs too late, the prior-favored interpretation has already become stable, making it difficult for local visual evidence to redirect the prediction trajectory.
To identify these critical locations, we quantitatively track changes in hidden-state predictions across network depth. Let the hidden state at the $l$-th Transformer layer be $\mathbf{h}^{(l)}\in\mathbb{R}^{d}$. Mapping it to the vocabulary $V$ through the language model head $\mathbf{W}_{\mathrm{lm}}$ yields the predictive distribution
$P^{(l)}=\mathrm{Softmax}(\mathbf{W}_{\mathrm{lm}}\mathbf{h}^{(l)})$. We use normalized predictive entropy to characterize the uncertainty of the prediction state at layer $l$:
\vspace{-3pt}
\begin{equation}
\mathcal{H}_{\mathrm{norm}}^{(l)}
=
-\frac{1}{\log |V|}
\sum_{v\in V}P^{(l)}(v)\log P^{(l)}(v).
\end{equation}
The entropy trajectory reflects three stages (Figure~\ref{fig:landscape}(b)).
First, the model aggregates multimodal evidence in a high-entropy state.
It then enters a stage of semantic decoding and prediction formation, during which the entropy curve deviates from its steady trend and exhibits pronounced nonlinear changes.
Finally, the model consolidates and refines candidate predictions, causing entropy to decrease to a low level and gradually converge.
Based on the geometric inflection points of the entropy curve, we use linear fitting to identify a set of critical layers $\mathcal{L}$ at which visual intervention is applied.
\subsection{Vertex-Centered Geometric Cues}
\begin{figure}[htbp]
  \centering
  \includegraphics[width=0.80\linewidth]{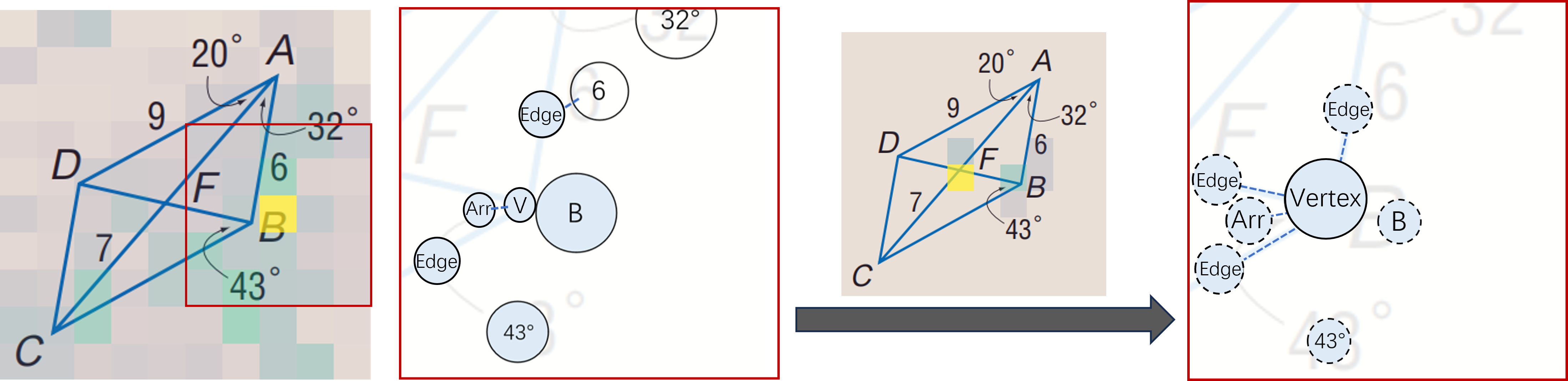}
  \caption{The Geometry-Constrained Local Relation Reconstruction (GCLR) mechanism.
  Using vertex $B$ as a spatial anchor, GCLR identifies its incident edges ($l_1$, $l_2$, $l_3$) and nearby geometric annotations.
  The resulting target token set and weight matrix explicitly compensate for the relative topological relations of the local structures.}
  \label{fig:gclr}
\end{figure}

Unlike natural images, angles, intersections, and collinearity relations in geometric diagrams are typically expressed jointly by discrete vertices, thin line segments, and nearby annotations. Such local topological information is visually subtle, yet crucial for correctly parsing geometric relations. During feature aggregation, attention tends to focus more readily on letters and numbers with explicit linguistic semantics, while weak edges, intersections, and their relative positional relations are easily overlooked.
Geometric vertices provide a natural organizing center for addressing this issue. Visually, edges, angles, intersections, and higher-order local configurations are organized around their incident vertices. Linguistically, geometric expressions such as $BD$ and $\angle ABC$ also refer to diagram entities through vertex labels. A vertex relevant to the current prediction can therefore be regarded as the center of a local semantic cluster: the visual representations surrounding it jointly encode the vertex label, incident edges, nearby annotations, and their topological relations (Figure~\ref{fig:gclr}). Because a vertex token covers a local image region and has been contextualized by the vision encoder, it can encode information about nearby incident edges and local structures. Preserving the salience of these vertex-centered representations enables subsequent FFN computations to organize nearby geometric evidence around the relevant vertices, thereby helping the model recover geometric relations such as edge connectivity, angle structures, and multi-edge intersections.
\vspace{-10pt}
\section{Method}
\begin{figure*}[htbp]
  \centering
  \includegraphics[width=0.85\textwidth]{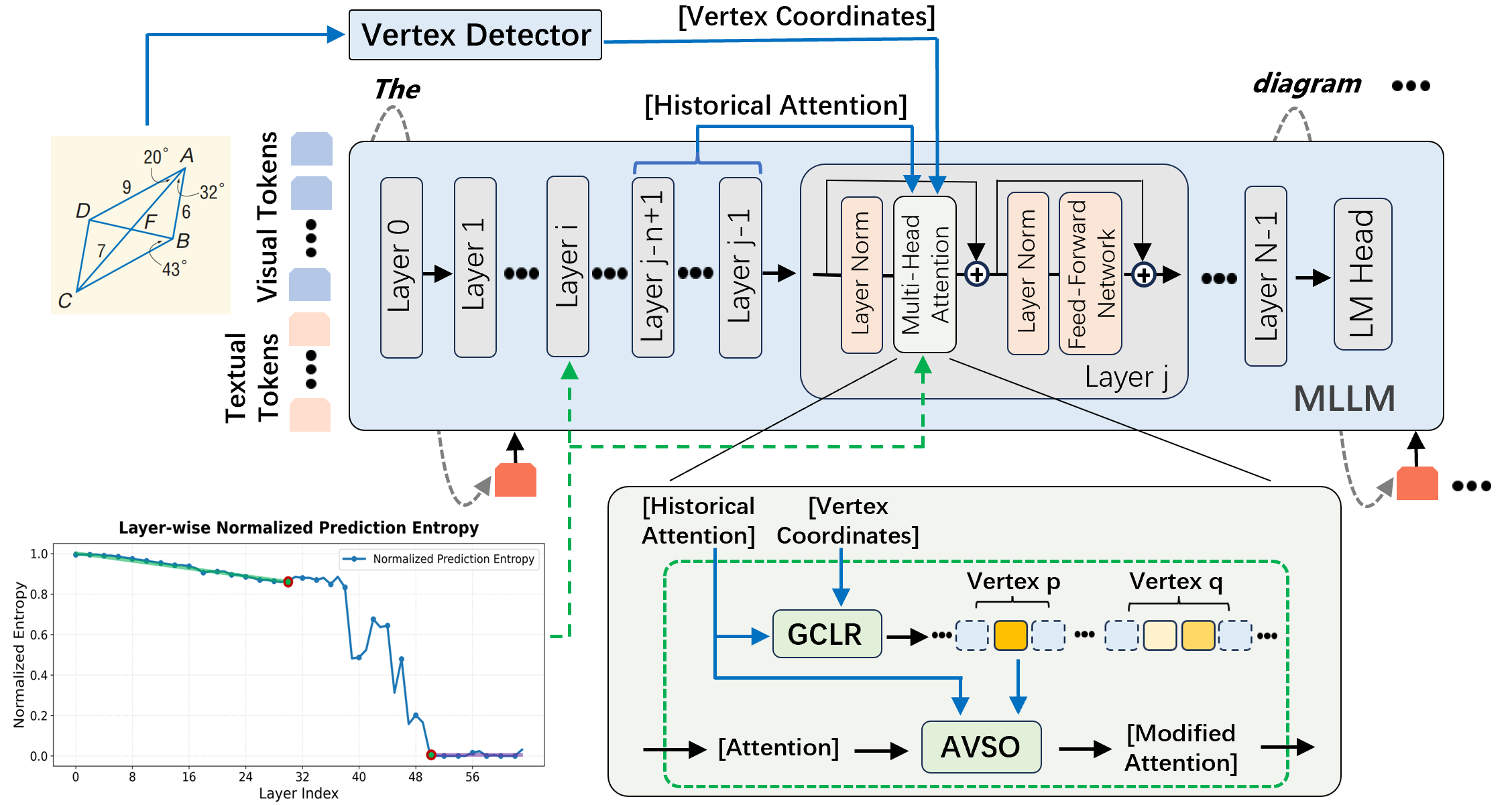} 
  \caption{Overview of CVIF. Layer-wise normalized predictive entropy provides a diagnostic for selecting candidate critical layers during prediction formation.  At these layers, GCLR utilizes geometric vertices as anchors to explicitly decouple and reconstruct spatial-semantic mappings for local structures. Subsequently, the AVSO performs distribution shaping and boost control over the attention, achieving active visual recall and prediction rectification.}
  \label{fig:cvif}
\end{figure*}
\subsection{Criticality-Driven Visual Intervention Framework}
As illustrated in Figure~\ref{fig:cvif}, the CVIF workflow proceeds as follows.
Given an image-text prompt, we identify a set of critical layers $\mathcal{L}$ based on the prediction entropy of the question-end token.
For each layer $l$, let $\mathbf{A}_{\text{curr}}^{(l)}\in\mathbb{R}^{H\times L_{\text{seq}}}$ denote the multi-head attention matrix, where $H$ is the number of heads and $L_{\text{seq}}$ is the sequence length.
We define $\bar{\mathbf{A}}\in\mathbb{R}^{L_{\text{seq}}}$ as the historical smoothed attention from the preceding $n$ layers (averaged across head dimensions), and $\mathcal{P}_{\text{vertices}}$ as the geometric vertex coordinates obtained from a vertex detector~\citep{yolov8_ultralytics}.
If $l\notin\mathcal{L}$, the model retains its native feed-forward path. Conversely, if $l\in\mathcal{L}$, the GCLR module leverages $\bar{\mathbf{A}}$ and $\mathcal{P}_{\text{vertices}}$ to compute a target token set $\Omega_{\text{tgt}}$ and a topological weight matrix $\mathbf{W}_{\text{target}}$:
\vspace{-3pt}
\begin{equation}
\Omega_{\text{tgt}},\mathbf{W}_{\text{target}}=\text{GCLR}(\bar{\mathbf{A}},\mathcal{P}_{\text{vertices}}).
\end{equation}
AVSO then safely reshapes the original attention $\mathbf{A}_{\text{curr}}^{(l)}$ within the multi-head attention module:
\vspace{-3pt}
\begin{equation}
\mathbf{A}_{\text{steered}}^{(l)}=\text{AVSO}(\mathbf{A}_{\text{curr}}^{(l)},\bar{\mathbf{A}},\Omega_{\text{tgt}},\mathbf{W}_{\text{target}}),
\end{equation}
The steered attention matrix $\mathbf{A}_{\text{steered}}$ is multiplied by the Value matrix, fusing key visual features to push the hidden state across the potential barrier in subsequent FFNs, steering subsequent representations toward visually supported predictions.
Finally, to ensure rigorous quantitative evaluation, the generated geometric descriptions are translated into formal language using Qwen3-VL-32B. 

\subsection{Geometry-Constrained Local Relation Reconstruction}

GCLR uses detected geometric vertices as anchors to select local visual evidence relevant to the current prediction. It first evaluates vertex relevance from historical attention and vertex-centered spatial priors, then identifies candidate tokens around selected vertices, and finally produces a target token set and topology-aware weights for subsequent attention steering.

Let $\mathcal{P}_{\text{vertices}}=
\{\mathbf{p}_k\}_{k=1}^{N_v}$ denote the detected vertex coordinates, where $N_v$ is the number of vertices. For each visual token $t\in\Omega_{\text{vis}}$, let $\mathbf{p}_t$ denote the center of its corresponding image region. The distance between vertex $k$ and token $t$ is defined as
$d_k(t)=\left\|\mathbf{p}_t-\mathbf{p}_k\right\|_2$.

For each vertex $k$, we construct a topology prior and an anchor prior:
$M_{\text{topo}}^{(k)}(t)=\rho\exp\left(-\frac{d_k(t)^2}{2\sigma_g^2}\right)
+(1-\rho)\exp\left(-\frac{d_k(t)}{\sigma_l}\right)$,
$M_{\text{anch}}^{(k)}(t)=\exp\left(-\frac{d_k(t)^2}{2\sigma_g^2}\right)$,
where $\rho=0.2$ controls the mixture of the two distance-decay components, and $\sigma_g$ and $\sigma_l$ control their spatial scales.
The base relevance of vertex $k$ is obtained through the weighted score of its historical attention under topological priors:
\vspace{-1pt}
\begin{equation}
s_{\text{base}}^{(k)}
=
\sum_{t\in\Omega_{\text{vis}}}
\bar{\mathbf{A}}(t)\,
m_{\text{topo}}^{(k)}(t).
\end{equation}
To incorporate distance-decay compensation, let
$d_k^{\text{ref}}$ denote the distance between vertex $k$ and the prediction-relevant reference location. The final vertex score is
\vspace{-3pt}
\begin{equation}
s_{\text{v}}^{(k)}
=
s_{\text{base}}^{(k)}
+
\eta
\max_{u\in\{1,\ldots,N_v\}}
s_{\text{base}}^{(u)}
\exp\left(-\frac{d_k^{\text{ref}}}{\tau}\right),
\end{equation}
where $\eta$ controls the compensation strength and $\tau$ controls its decay rate.
We retain vertices whose scores exceed a dynamic threshold:
$
\mathcal{V}_{\text{act}}
=
\left\{
k \,\middle|\,
s_{\text{v}}^{(k)}
\geq
\mu_{\text{v}}+\alpha_{\text{v}}\sigma_{\text{v}}
\right\},
$
where $\mu_{\text{v}}$ and $\sigma_{\text{v}}$ are the mean and standard deviation of the candidate vertex scores, respectively. The normalized vertex weight is
$
\omega_{\text{v}}^{(k)}
=
\frac{s_{\text{v}}^{(k)}}
{\sum_{u\in\mathcal{V}_{\text{act}}}s_{\text{v}}^{(u)}},
k\in\mathcal{V}_{\text{act}}.
$
For each activated vertex $k$, the token-level response is
\vspace{-3pt}
\begin{equation}
s_{\text{tok}}^{(k)}(t)
=
M_{\text{topo}}^{(k)}(t)
\bar{\mathbf{A}}(t),
\qquad
t\in\Omega_{\text{vis}}.
\end{equation}
After applying the dynamic threshold
$\mu_{\text{tok}}^{(k)}+
\alpha_{\text{tok}}\sigma_{\text{tok}}^{(k)}$,
we retain the representative tokens as
$\mathcal{C}_{\text{feature}}^{(k)}$.
To preserve the vertex-centered anchor, we additionally select
$
c_k
=
\arg\max_{t\in\Omega_{\text{vis}}}
M_{\text{anch}}^{(k)}(t)
\bar{\mathbf{A}}(t).
$
The final candidate set for vertex $k$ is
$
\mathcal{U}_k
=
\mathcal{C}_{\text{feature}}^{(k)}
\cup
\{c_k\}.
$
Within $\mathcal{U}_k$, we combine anchor localization and historical attention to compute token-level weights:
\vspace{-3pt}
\begin{equation}
\omega_{\text{tok}}^{(k,j)}
=
(1-\gamma)
\frac{m_{\text{anch}}^{(k)}(j)}
{\sum_{u\in\mathcal{U}_k}m_{\text{anch}}^{(k)}(u)}
+
\gamma
\frac{\bar{\mathbf{A}}(j)}
{\sum_{u\in\mathcal{U}_k}\bar{\mathbf{A}}(u)},
\qquad
j\in\mathcal{U}_k,
\end{equation}
where $\gamma$ balances spatial anchoring and historical attention.
The final target token set is
$
\Omega_{\text{tgt}}
=
\bigcup_{k\in\mathcal{V}_{\text{act}}}
\mathcal{U}_k.
$
For each target token $t\in\Omega_{\text{tgt}}$, we aggregate its contributions from all activated vertices:
\vspace{-3pt}
\begin{equation}
\mathbf{W}_{\text{target}}(t)
=
\sum_{k\in\mathcal{V}_{\text{act}}}
\omega_{\text{v}}^{(k)}
\omega_{\text{tok}}^{(k,t)}
\mathbb{I}[t\in\mathcal{U}_k].
\end{equation}

\subsection{Adaptive Visual Steering Operator}

\begin{figure*}[htbp]
  \centering
  \includegraphics[width=0.75\textwidth]{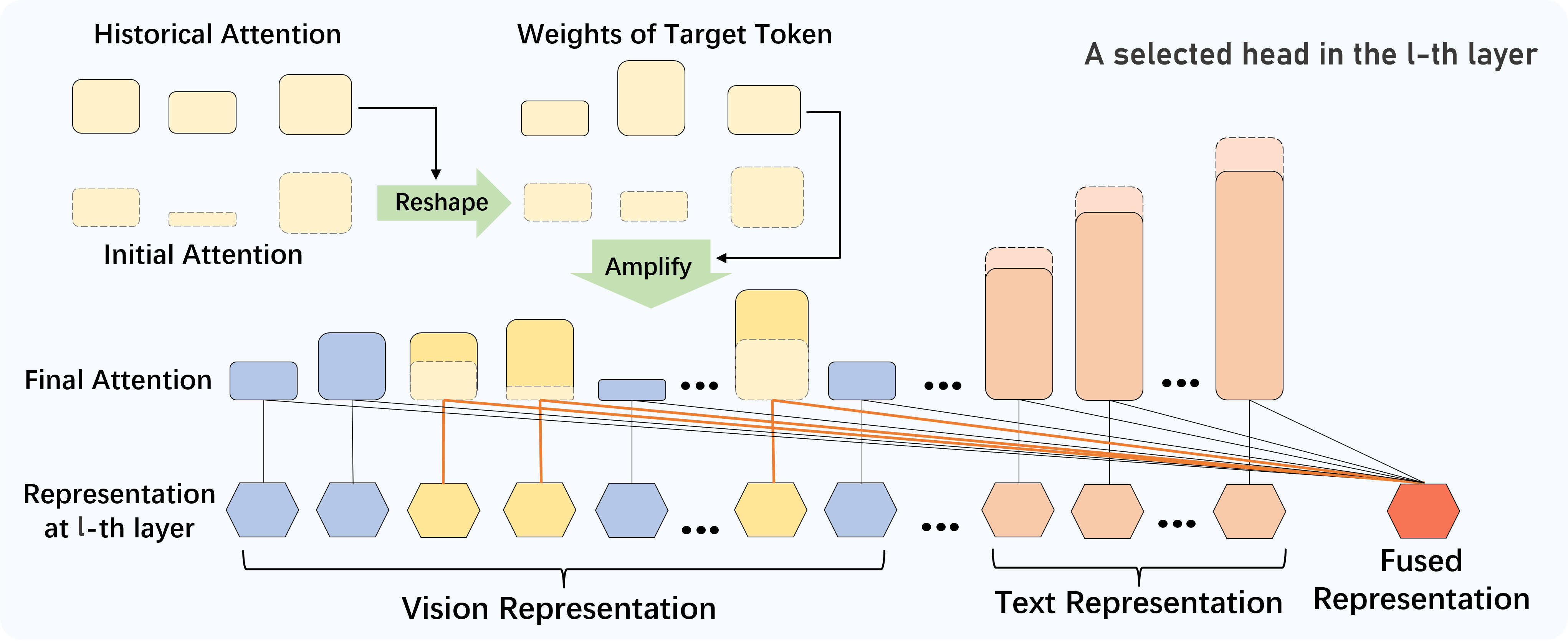}
  \caption{The principle of the Adaptive Visual Steering Operator (AVSO).
}
\end{figure*}

As illustrated in Figure 5, AVSO reshapes attention at critical layers without modifying model parameters. It uses the target-token set and topology-aware weights produced by GCLR, together with historical and current attention, to increase the influence of key geometric evidence while compensating for the added attention mass through text tokens.

For each target token $t\in\Omega_{\text{tgt}}$,
$\mathbf{A}_{\text{curr}}^{(l)}(t)$ denotes its current attention values across heads,
$\bar{\mathbf{A}}(t)$ denotes its head-aggregated historical attention prior, and
$\mathbf{W}_{\text{target}}(t)$ denotes its normalized topology-aware weight.
All vector operations below are applied element-wise along the head dimension.
AVSO first compares current attention with the historical attention prior to compute a shaping factor for each target token:
\vspace{-3pt}
\begin{equation}
\mathbf{M}_{\text{shape}}^{(l)}(t)
=
(1-\alpha)\mathbf{1}
+
\alpha
\frac{\bar{\mathbf{A}}(t)}
{\mathbf{A}_{\text{curr}}^{(l)}(t)+\epsilon},
\qquad
t\in\Omega_{\text{tgt}},
\end{equation}
where $\alpha\in[0,1]$ controls the contribution of historical attention, $\epsilon$ is a numerical stability constant, and $\mathbf{1}$ is an all-one vector over attention heads.
The shaped target attention is then defined as
\vspace{-3pt}
\begin{equation}
\mathbf{A}_{\text{shaped}}^{(l)}(t)
=
\mathbf{A}_{\text{curr}}^{(l)}(t)
\odot
\mathbf{M}_{\text{shape}}^{(l)}(t),
\qquad
t\in\Omega_{\text{tgt}},
\end{equation}
where $\odot$ denotes element-wise multiplication.
AVSO selectively triggers intervention only for heads exhibiting high attention toward visual tokens yet insufficient alignment with the target tokens.
\vspace{-3pt}
\begin{equation}
\boldsymbol{\Phi}_{\text{gate}}^{(l)}
=
\mathbb{I}\left(
\sum_{t\in\Omega_{\text{vis}}}
\mathbf{A}_{\text{curr}}^{(l)}(t)
\geq
\tau_{\text{down}}
\;\land\;
\sum_{t\in\Omega_{\text{tgt}}}
\mathbf{A}_{\text{curr}}^{(l)}(t)
\leq
\tau_{\text{up}}
\right),
\end{equation}
where $\tau_{\text{down}}$ and $\tau_{\text{up}}$ are the lower visual-attention threshold and upper target-attention threshold, respectively.
To ensure that the shaped target attention reaches a competitive visual-attention scale, we compute a guarantee coefficient:
$
\boldsymbol{\gamma}_{\text{guarantee}}^{(l)}
=
\operatorname{Clamp}\left(
\frac{
\max_{t\in\Omega_{\text{vis}}}
\mathbf{A}_{\text{curr}}^{(l)}(t)
}{
\min_{t\in\Omega_{\text{tgt}}}
\mathbf{A}_{\text{shaped}}^{(l)}(t)
+\epsilon
},
\,1.0,\,+\infty
\right).
$
For each target token, AVSO updates its attention as
\vspace{-3pt}
\begin{equation}
\resizebox{0.9\linewidth}{!}{$
\mathbf{A}_{\text{boosted}}^{(l)}(t)
=
\mathbf{A}_{\text{curr}}^{(l)}(t)
\odot
\left[
\boldsymbol{\Phi}_{\text{gate}}^{(l)}
\odot
\mathbf{M}_{\text{shape}}^{(l)}(t)
\odot
\boldsymbol{\gamma}_{\text{guarantee}}^{(l)}
\cdot
\lambda
\cdot
\mathbf{W}_{\text{target}}(t)
+
\left(
\mathbf{1}
-
\boldsymbol{\Phi}_{\text{gate}}^{(l)}
\right)
\right],
t\in\Omega_{\text{tgt}},
$}
\end{equation}
where $\lambda>0$ controls the intervention strength.
The attention mass added to target tokens is
$
\Delta\mathbf{P}^{(l)}
=
\sum_{t\in\Omega_{\text{tgt}}}
\left(
\mathbf{A}_{\text{boosted}}^{(l)}(t)
-
\mathbf{A}_{\text{curr}}^{(l)}(t)
\right).
$
Let the current attention mass assigned to text tokens be
$
\mathbf{m}_{\text{text}}^{(l)}
=
\sum_{t\in\Omega_{\text{text}}}
\mathbf{A}_{\text{curr}}^{(l)}(t).
$
The text-token scaling factor is defined as
$
\boldsymbol{\beta}_{\text{text}}^{(l)}
=
\operatorname{Clamp}\left(
\mathbf{1}
-
\frac{\Delta\mathbf{P}^{(l)}}
{\mathbf{m}_{\text{text}}^{(l)}+\epsilon},
\,0,\,1
\right).
$
The compensated text attention is
\vspace{-3pt}
\begin{equation}
\mathbf{A}_{\text{text}}^{(l)}(t)
=
\mathbf{A}_{\text{curr}}^{(l)}(t)
\odot
\boldsymbol{\beta}_{\text{text}}^{(l)},
\qquad
t\in\Omega_{\text{text}}.
\end{equation}
Combining the boosted attention of target tokens, the compensated attention of text tokens, and the unchanged attention of all remaining tokens, followed by normalization over the token dimension, yields the final steered attention $\mathbf{A}_{\text{steered}}^{(l)}$.

\section{Experiment}
\vspace{-3pt}
\subsection{Experimental Setup}
We evaluated the performance of geometric diagram parsing on 2,650 image–formal language pairs from PGPS9K~\citep{zhang2023pgps9k} and 2,650 pairs from PGDP5K~\citep{Hao2022PGDP5KAD}. To enable unified evaluation, we convert PGDP5K annotations into the PGPS representation. By comparing the predicted set with the ground-truth set, we report precision (P), recall (R), and F1 for structural, semantic, and overall predicates.
CVIF uses Qwen3-VL-32B-Instruct~\citep{bai2025qwen3} as the intervention backbone. We compare it with open-source models from the Qwen, InternVL, and GLM families~\citep{qwen2.5vl,bai2025qwen3,wang2025internvl3,qwen35omni2026,vteam2025glm45v}, as well as proprietary models~\citep{claude45,openai2025gpt5,google2025gemini3}.
\vspace{-8pt}
\begin{table*}[htbp]
\centering
\caption{Performance comparison on PGPS9K.}
\label{tab:parsing}
\resizebox{\textwidth}{!}{
\begin{tabular}{l|ccc|ccc|ccc}
\toprule
\multirow{2}{*}{Methods} & \multicolumn{3}{c|}{Structure} & \multicolumn{3}{c|}{Semantic} & \multicolumn{3}{c}{Overall} \\
\cline{2-10}
 & P & R & F1 & P & R & F1 & P & R & F1 \\
\midrule
\multicolumn{10}{c}{\textit{\textbf{Open-source MLLMs}}} \\
Qwen2.5-VL-7B-Instruct  & 73.58 & 38.00 & 50.12 & 38.32 & 43.82 & 40.89 & 47.19 & 40.70 & 43.70 \\
InternVL3.5-8B-Instruct & 66.14 & 43.70 & 52.63 & 43.42 & 51.55 & 47.14 & 51.44 & 47.38 & 49.33 \\
InternVL3.5-30B-Instruct& 70.03 & 58.89 & 63.98 & 57.39 & 56.12 & 56.75 & 62.32 & 57.61 & 59.87 \\
InternVL3.5-8B        & 68.90 & 54.26 & 60.71 & 61.52 & 59.06 & 60.26 & 64.00 & 56.49 & 60.01 \\
Qwen2.5-VL-32B-Instruct & 70.15 & 53.01 & 60.39 & 63.33 & 58.9  & 61.03 & 65.49 & 55.74 & 60.22 \\
Qwen3-VL-8B-Instruct    & 72.75 & 52.12 & 60.73 & 73.56 & 66.82 & 70.03 & 71.18 & 58.92 & 64.47 \\
GLM-4.6V-Flash  & 77.25 & 52.08 & 62.22 & 77.39 & 62.68 & 69.26 & 76.42 & 56.98 & 65.29 \\
Qwen3.5-9B  & 80.45 & 66.94 & 73.08 & 78.38 & 75.87 & 77.10 & 77.60 & 71.08 & 74.19 \\
Qwen3-VL-32B-Instruct   & 82.91 & 76.16 & 79.39 & 79.63 & 74.67 & 77.07 & 80.37 & 75.48 & 77.85 \\
\midrule
\multicolumn{10}{c}{\textit{\textbf{Closed-source MLLMs}}} \\
Claude-4.5-Sonnet       & 81.47 & 69.25 & 74.87 & 78.17 & 68.93 & 73.26 & 79.04 & 69.10 & 73.74 \\
GPT-5.2-1211            & 95.52 & 90.66 & 93.03 & 88.93 & 84.68 & 86.75 & 91.98 & 87.90 & 89.89 \\
Gemini-3-Flash         & 98.84 & 91.14 & 94.84 & 86.73 & 84.14 & 85.41 & 92.91 & 87.90 & 90.33 \\
\midrule
CVIF (Ours)   & 90.73 & 88.41 & 89.56 & 83.28 & 81.95 & 82.61 & 85.72 & 85.43 & 85.58 \\
\bottomrule
\end{tabular}
}
\end{table*}
\vspace{-10pt}
\subsection{Diagram Parsing Results}
CVIF improves Overall F1 over Qwen3-VL-32B-Instruct from 77.85 to 85.58 on PGPS9K and from 75.23 to 82.84 on PGDP5K (Tables~\ref{tab:parsing} and~\ref{tab:pgdp}).
On PGPS9K, structural and semantic F1 improve by 10.17 and 5.54 points, respectively; the corresponding gains on PGDP5K are 9.77 and 5.95 points.
CVIF also consistently improves overall precision and recall: by 5.35 and 9.95 points on PGPS9K, and by 4.84 and 10.15 points on PGDP5K.
The larger recall gains indicate more complete recovery of geometric predicates, while the simultaneous precision gains show that this improved coverage does not introduce additional spurious relations.
CVIF achieves the best Overall F1 among the evaluated open-source models on both datasets.
\vspace{-10pt}
\begin{table*}[htbp]
\centering
\caption{Performance comparison on PGDP5K.}
\label{tab:pgdp}
\resizebox{\textwidth}{!}{
\begin{tabular}{l|ccc|ccc|ccc}
\toprule
\multirow{2}{*}{Methods} & \multicolumn{3}{c|}{Structure} & \multicolumn{3}{c|}{Semantic} & \multicolumn{3}{c}{Overall} \\
\cline{2-10}
 & P & R & F1 & P & R & F1 & P & R & F1 \\
\midrule
Qwen2.5-VL-32B-Instruct & 70.15 & 53.01 & 60.39 & 63.33 & 58.9  & 61.03 & 65.49 & 55.74 & 60.22 \\
Qwen3-VL-8B-Instruct     & 73.16 & 51.04 & 60.13 & 71.15 & 60.15 & 65.19 & 69.89 & 55.15 & 61.65 \\
GLM-4.6V-Flash   & 76.95 & 53.18 & 62.89 & 71.27 & 56.73 & 63.18 & 72.04 & 54.78 & 62.24 \\
Qwen3.5-9B   & 79.72 & 65.21 & 71.74 & 77.98 & 72.45 & 75.11 & 76.70 & 68.48 & 72.36 \\
Qwen3-VL-32B-Instruct   & 81.28 & 74.58 & 77.79 & 77.43 & 69.81 & 73.43 & 78.26 & 72.43 & 75.23 \\
GPT-5.2-1211            & 94.03 & 89.47 & 91.69 & 84.50 & 79.67 & 82.01 & 89.20 & 85.04 & 87.07 \\
Gemini-3-Flash          & 97.15 & 93.54 & 95.31 & 77.62 & 75.67 & 76.63 & 88.30 & 85.47 & 86.86 \\
\midrule
CVIF (Ours)   & 88.76 & 86.38 & 87.56 & 80.84 & 77.97 & 79.38 & 83.10 & 82.58 & 82.84 \\
\bottomrule
\end{tabular}
}
    \vspace{-10pt}
\end{table*}

\begin{table*}[h]
    \vspace{-10pt}
\centering
\caption{Sample Accuracy and Perfect Parsing Rate (PPR) on PGPS9K and PGDP5K.}
\label{tab:exact_match}
\resizebox{\textwidth}{!}{
\small
\begin{tabular}{l|ccccc|ccccc}
    \toprule
\multirow{2}{*}{Methods} & \multicolumn{5}{c|}{PGPS9K} & \multicolumn{5}{c}{PGDP5K} \\
\cline{2-11}
& Lines & Circles & Structures & Semantics & PPR & Lines & Circles & Structures & Semantics & PPR \\
\midrule
Qwen2.5-VL-32B-Instruct & 36.57 & 77.51 & 34.75 & 38.42 & 21.43 & 34.64 & 78.64 & 32.60 & 38.42 & 19.36 \\
Qwen3-VL-8B-Instruct & 42.79 & 77.51 & 38.08 & 50.75 & 25.06 & 42.45 & 77.85 & 37.40 & 50.00 & 24.00 \\
GLM-4.6V-Flash & 43.70 & 80.60 & 39.09 & 49.81 & 25.77 & 46.34 & 80.00 & 39.70 & 51.02 & 25.40 \\
Qwen3.5-9B & 56.00 & 85.43 & 50.64 & 61.89 & 37.36 & 55.55 & 82.60 & 48.30 & 63.47 & 36.34 \\
Qwen3-VL-32B-Instruct & 64.23 & 92.57 & 62.38 & 58.79 & 43.47 & 64.38 & 87.62 & 58.08 & 58.79 & 40.11 \\
GPT-5.2-1211 & 84.42 & 91.36 & 79.13 & 73.55 & 62.04 & 85.81 & 86.42 & 75.66 & 71.40 & 58.26 \\
Gemini-3-Flash & 89.32 & 91.77 & 83.92 & 77.21 & 67.77 & 94.23 & 89.85 & 85.02 & 68.15 & 60.68 \\
\midrule
CVIF (Ours) & 79.09 & 95.62 & 77.25 & 69.96 & 59.47 & 80.00 & 88.72 & 72.23 & 69.77 & 55.77 \\
\bottomrule
\end{tabular}
}
    \vspace{-10pt}
\end{table*}
\subsection{Diagram-level Exact Match Evaluation}
Category-level F1 does not guarantee that an entire diagram is parsed correctly. We therefore report Sample Accuracy (SA), which requires all predicates within a category to match the ground truth, and Perfect Parsing Rate (PPR), which requires all predicates across the diagram to be correct.
As shown in Table~\ref{tab:exact_match}, CVIF achieves the best results among the evaluated open-source configurations on all metrics.
Compared with Qwen3-VL-32B-Instruct, CVIF raises PPR from 43.47 to 59.47 on PGPS9K and from 40.11 to 55.77 on PGDP5K.
These results show that CVIF improves not only individual predicate recovery, but also the global completeness and correctness of diagram parsing.
\begin{table*}[htbp]
    \vspace{-10pt}
\centering
\caption{Ablation results on PGPS9K.}
\label{tab:ablation}
\resizebox{\textwidth}{!}{
\begin{tabular}{l|ccc|ccc|ccc}
\toprule
\multirow{2}{*}{Methods} & \multicolumn{3}{c|}{Structure} & \multicolumn{3}{c|}{Semantic} & \multicolumn{3}{c}{Overall} \\
\cline{2-10}
 & P & R & F1 & P & R & F1 & P & R & F1 \\
\midrule
Qwen3-VL-32B (Baseline) & 82.91 & 76.16 & 79.39 & 79.63 & 74.67 & 77.07 & 80.37 & 75.48 & 77.85 \\
Qwen3-VL-32B 2-stage     & 84.29 & 78.92 & 81.52 & 81.11 & 77.85 & 79.45 & 81.11 & 77.85 & 79.45 \\
Uniform Intv.  & 86.00 & 80.13 & 82.96 & 79.66 & 77.03 & 78.32 & 81.07 & 78.70 & 79.86 \\
w/o GCLR     & 86.46 & 81.21 & 83.75 & 81.49 & 78.79 & 80.12 & 82.51 & 80.10 & 81.28 \\
w/o AVSO     & 86.12 & 81.11 & 83.54 & 80.34 & 78.54 & 79.43 & 81.12 & 79.92 & 80.52 \\
\midrule
CVIF (Ours)  & 90.73 & 88.41 & 89.56 & 83.28 & 81.95 & 82.61 & 85.72 & 85.43 & 85.58 \\
\bottomrule
\end{tabular}
}
    \vspace{-10pt}
\end{table*}
\vspace{-10pt}
\subsection{Ablation Study}
\vspace{-5pt}
Table~\ref{tab:ablation} shows that the two-stage pipeline, which generates a natural-language description before formalizing it, reaches 79.45 Overall F1, substantially below CVIF (85.58). Thus, CVIF's gains cannot be explained by changing the output pipeline alone. ``Uniform Intv.'' replaces criticality-guided layer selection with fixed interventions at layers 16, 32, and 48 of the 64-layer backbone, yielding 79.86 Overall F1 versus 85.58 for CVIF and supporting criticality-guided placement. Replacing GCLR with Top-$K$ visual-token selection reduces Overall F1 to 81.28, while replacing AVSO with direct attention amplification reduces it to 80.52. These results verify the importance of both criticality-guided layer selection and the two proposed modules.

\begin{figure*}[t]
\centering
\includegraphics[width=\textwidth]{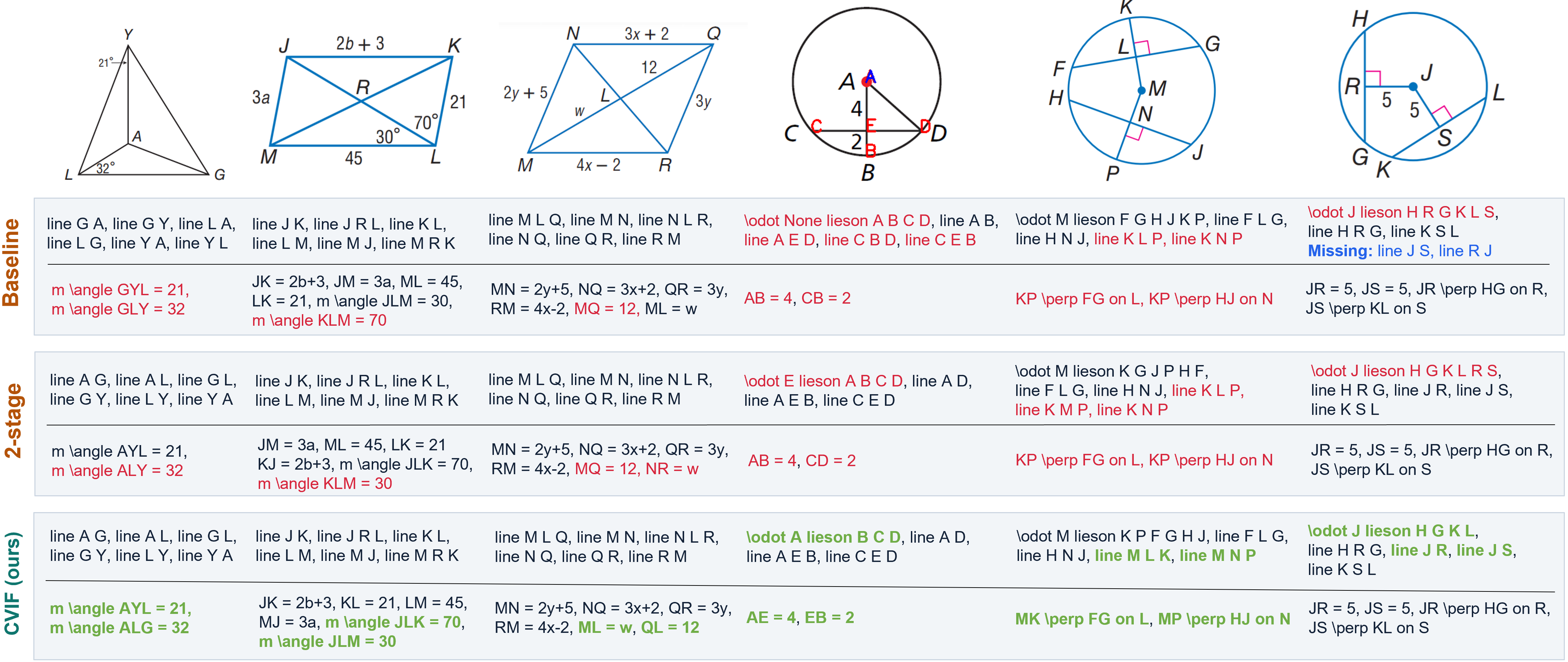}
\caption{Qualitative case analysis on representative geometric diagrams. }
\label{fig:case_analysis}
\vspace{-10pt}
\end{figure*}
\subsection{Case Analysis}
Figure~\ref{fig:case_analysis} compares the parsing results of the baseline, the two-stage pipeline, and CVIF on representative diagrams. Although the baseline and two-stage pipeline recover many individual predicates, they often assign attributes to incorrect geometric entities or miss local structural relations required for correct parsing. In contrast, CVIF consistently recovers the relevant vertex-centered topology, including angle assignments, line connectivity, collinearity, and perpendicular relations. These examples illustrate that the proposed visual intervention helps the model retain and utilize local geometric evidence, thereby producing structurally consistent diagram parses.
\vspace{-5pt}
\subsection{Downstream Geometry Reasoning}
\vspace{-5pt}
We evaluate the downstream utility of diagram representations on the Vision Dominant split of the Plane Geometry subset of MathVerse~\citep{zhang2024mathverse}. This representative and challenging geometry-reasoning setting requires models to rely primarily on visual evidence, making it well suited to evaluate whether improved diagram perception transfers to downstream reasoning. We compare five reasoning models under three settings: direct answering, augmentation with a formal-language representation, and augmentation with a natural-language diagram description (Table~\ref{tab:downstream}). Natural-language descriptions improve accuracy for all five models, with gains from 1.76 to 5.88 points. Formal-language representations improve four models, while the preferred representation varies by backbone. These results show that explicit diagram representations can improve downstream geometry reasoning in visually demanding settings.
\vspace{-10pt}
\begin{table*}[h]
\centering
\caption{Downstream reasoning accuracy on MathVerse.}
\label{tab:downstream}
\resizebox{\textwidth}{!}{%
\begin{tabular}{lccccc}
\toprule
Methods & Qwen3-VL-32B & Qwen3-VL-8B & InternVL3.5-8B & Qwen2.5-VL-7B & Qwen3-VL-4B \\
\midrule
Direct Answer & 53.73 & 45.10 & 41.18 & 42.16 & 41.18 \\
+ Formal Language & 53.73 & 49.61 & 45.29 & 46.08 & 41.37 \\
+ Text Description & 58.63 & 48.24 & 47.06 & 46.47 & 42.94 \\
\bottomrule
\end{tabular}%
}
    \vspace{-10pt}
\end{table*}
\vspace{-10pt}
\section{Conclusion}
\vspace{-5pt}
In this paper, we proposed a training-free Criticality-Driven Visual Intervention Framework to mitigate textual-prior bias in MLLMs' geometric diagram understanding.
An energy-landscape perspective motivates reinforcing local visual evidence at representational critical states, where competing interpretations have not yet become stable. The framework uses layer-wise prediction diagnostics to select candidate critical layers and directs attention toward vertex-guided visual evidence.
We designed the GCLR, which uses geometric vertices as anchors to reconstruct "spatial-semantic" mappings, overcoming the failure of global attention to capture complex topologies. Furthermore, by introducing the AVSO, our framework achieved robust distribution reshaping.
The reported experiments show improved diagram-parsing performance with the intervention, and evaluate the use of generated descriptions in downstream geometry reasoning, demonstrating a novel inference-time intervention paradigm for mitigating visual hallucinations.

\bibliography{references}
\bibliographystyle{iclr2026_conference}

\end{document}